# Local Path Planning with Dynamic Obstacle Avoidance in Unstructured Environments

Okan Arif Guvenkaya[1]    Selim Ahmet Iz[2]    Mustafa Unel[1]
[1]*Faculty of Engineering and Natural Sciences, Sabanci University,* Istanbul, Turkey
[2]*Institute of Optical Sensor Systems, German Aerospace Center (DLR),* Berlin, Germany
selim.iz@dlr.de    {okanarif, munel}@sabanciuniv.edu

*Abstract*—Obstacle avoidance and path planning are essential for guiding unmanned ground vehicles (UGVs) through environments that are densely populated with dynamic obstacles. This paper develops a novel approach that combines tangent-based path planning and extrapolation methods to create a new decision-making algorithm for local path planning. In the assumed scenario, a UGV has a prior knowledge of its initial and target points within the dynamic environment. A global path has already been computed, and the robot is provided with waypoints along this path. As the UGV travels between these waypoints, the algorithm aims to avoid collisions with dynamic obstacles. These obstacles follow polynomial trajectories, with their initial positions randomized in the local map and velocities randomized between 0 and the allowable physical velocity limit of the robot, along with some random accelerations. The developed algorithm is tested in several scenarios where many dynamic obstacles move randomly in the environment. Simulation results show the effectiveness of the proposed local path planning strategy by gradually generating a collision free path which allows the robot to navigate safely between initial and the target locations.

## I. INTRODUCTION

Autonomous vehicles, a significant point of research, are in high demand for navigating human environments and performing diverse tasks like self-driving cars, delivery drones, and service robots [1], [2]. Path planning, vital for safe navigation in complex environments, stands out as a major challenge for these systems.

Path planning algorithms are typically classified based on several criteria, including the nature of obstacles (static or dynamic), the planning approach (global or local), and the environmental conditions (known or unknown). Global path planning proves effective in static and known environments, leveraging extensive datasets to chart a course. Conversely, local path planning becomes required in dynamic or unknown environments, where robots rely on local sensors to adapt their trajectory as the environment evolves, offering a more adaptable approach compared to global planning [3] [4]. Highly dynamic environments, characterized by multiple static and dynamic obstacles, pose additional complexities, amplified by factors such as the behavior of dynamic obstacles, the strategic placement of surveillance sensors, and onboard sensor ranges [5] [6] [7].

Path planning algorithms cover various approaches, each address distinct challenges across varied environments. Graph-based algorithms like Dijkstra's, A*, and D* Lite excel in static landscapes with known obstacles, while sampling-based methods such as Probabilistic Roadmaps (PRM) and Rapidly-exploring Random Trees (RRT) succeed in dynamic environments by probing random points. Heuristic algorithms, including Potential Fields, Genetic Algorithms, and Simulated Annealing, offer intuitive approximations for efficient path navigation, even though potentially sacrificing optimality. Furthermore, hybrid algorithms, combining different techniques, provide a comprehensive toolkit for adaptable pathfinding across diverse terrains [8].

Various studies have explored innovative approaches to path planning in dynamic environments. Conflict-based Search (CBS) in conjunction with the D* Lite algorithm demonstrates efficacy in navigating unknown dynamic environments by harmonizing individual robot path planning with collision avoidance for multiple robots [9]. Integration of Heuristic Search Based Algorithms like A* with Potential Fields presents a strategy for navigating Unmanned Surface Vehicles (USVs) through dynamic environments, combining global path planning with real-time obstacle avoidance [10]. Advancements such as RRT, RRT*, and Improved Bidirectional RRT* highlight efficient path planning methods for smart vehicles in dynamic settings, integrating vehicle-specific constraints and collision detection mechanisms [11]. Collaboration between Ant Colony Optimization (ACO) and the Dynamic Window Approach (DWA) enables effective multi-robot navigation and obstacle avoidance within complex terrains, leveraging globally optimized paths generated by ACO and real-time obstacle avoidance by DWA [12]. Additionally, approaches like Maximum-Speed Aware Velocity Obstacle (MVO) ensure safe navigation in the presence of high-speed obstacles, showcasing effectiveness in collision avoidance [13]. Moreover, deep learning algorithms, exemplified by ANOA (Autonomous Navigation and Obstacle Avoidance) using Q-learning, offer autonomous navigation capabilities superior to conventional methods in static and dynamic environments [13].

This paper develops a novel local path planning algorithm tailored for highly dynamic and complex environments where the map is initially unknown. An initial global map can be constructed with the help of other robots, such as UAVs, or with alternative methods. Using this global map information, global path planning is completed with any well-known algorithm, such as A* or RRT*. Afterwards, extracted waypoints are provided to the robot. The algorithm proposed in this paper focuses on traveling between waypoints. It relies on two key environmental data which are environmental density and real-time detection of dynamic obstacles. The environmental density, which defines the density of moving obstacles across the drivable terrain, is acquired during the initial global path construction. Real-time detection of dynamic obstacles can be facilitated by UAVs or onboard sensors such as LIDAR or cameras. Utilizing the global map information, the UGV navigates using the proposed local path planning algorithm to circumvent collisions with dynamic obstacles in complex environments. The primary contribution of this paper is introduction of a novel decision algorithm for local path planning, aimed at avoiding dynamic obstacles in highly dynamic environments to enhance path safety and reduce travel time. This approach draws inspiration from tangent-based methods and the dynamic window approach, supported by rigorous numerical analysis incorporating future state estimation techniques for dynamic obstacles. As detailed in the methodology section, the proposed algorithm does not consider obstacles that are outside the critical area, and thus, computational cost is decreased.

The paper is structured as follows: Section 2 provides a detailed description of the methodology used in this work. Section 3 presents and discusses some simulation results. Finally, Section 4 concludes the paper with some remarks and indicate possible future directions.

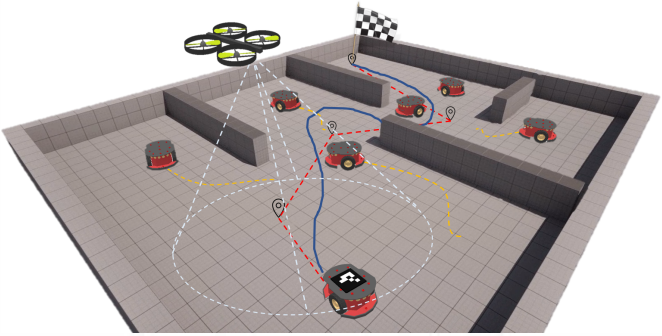

Fig. 1. Environment visualisation with CoppeliaSim.

## II. METHODOLOGY

Continuous tracking and surveillance is one of the most essential steps of the all dynamic obstacle avoidance algorithms. This surveillance process can be done by using different types of onboard sensors or creating a collaborative system where there exist different assistant robots besides master. UAVs are widely preferred assistant robots for collaborative studies because of their wide looking angles and high degree of freedoms [14]. They can easily adopt to mosaicking whole maps [15], or help to find the proper routes even by considering the structure of the terrain [16] besides surveillance of dynamic scenes. As it can be seen in the Fig. 1, the master robot, which have a QR on top of it, tries to reach to desired location which represented by a flag, by avoiding yellow routes that represent the predicted routes of the dynamic obstacles. Additionally, the blue route shows the path UGV followed by avoiding from the possible collisions.

The environment is highly dynamic and complex. The global waypoints are predetermined by one of the methods mentioned in the introduction. The robot needs to follow these waypoints sequentially. For each travel from one waypoint to another, our proposed local path planning algorithm is used.

The initial position of the robot is $(x_{\text{initial}}, y_{\text{initial}})$, and the aimed waypoint is $(x_{\text{target}}, y_{\text{target}})$. The map area $(A_{\text{map}})$ can be calculated as:

$$A_{\text{map}} = |x_{\text{target}} - x_{\text{initial}}| \times |y_{\text{target}} - y_{\text{initial}}| \quad (1)$$

The velocity of the robot is $v$ [m/s] where $v \in [0, V]$. $V$ is the maximum velocity limit of the robot. The existing literature offers numerous approaches to adjusting the velocity of robots, but the proposed approach introduces a new local path planning decision. This decision aims to find the optimum sensing region while maintaining a constant and high velocity for the robot to decrease travel time while ensuring high safety. [20]

There are $n$ dynamic obstacles, in the area where $n \in [0, N]$. $N$ is the maximum possible dynamic obstacle in the map. Each dynamic obstacle can reach a maximum velocity the same as the robot, $V$ [m/s]. The robot can search a closer area with the help of another helper robot, such as a UAV, or with its onboard sensors. If onboard sensors are used, then the maximum range is $r_{\text{sensing}}^{\text{max}}$ [m], which represents the furthest distance the sensor can read. Consequently, the sensing region of the robot forms a circle with a radius of $r_{\text{sensing}}^{\text{max}}$. Otherwise, the sensing region is limited by the maximum altitude of the UAV and its camera qualifications.

The dynamic obstacles randomly travel around the map, and they have no intention of whether to collide or not collide with the robot. They exhibit random accelerations, and they may follow higher degree polynomial trajectories.

For simplicity of discussion, let's assume that the robot and dynamic obstacles are circular in shape, with radii of $r_{\text{robot}}$ and $r_{\text{obstacle}}$ respectively.

The complexity of the dynamic environment is correlated with the obstacle density $(\rho_{\text{obstacle}})$ in the drivable area, which is calculated as follows.

$$\rho_{\text{obstacle}} = \frac{n}{A_{\text{map}}} \quad (2)$$

### A. Proposed Dynamic Obstacle Avoidance Approach via Extrapolation

In the algorithm, several key terms are defined, crucial for navigation and obstacle avoidance. Their visual representation can be seen in Fig. 2. The critical area is an imaginary circular zone around the robot, monitored by its sensors, with a radius $(r_{\text{ca}})$ ranging from $r_{\text{robot}}$ to $r_{\text{sensing}}^{\text{max}}$. The dynamic obstacle safe zone, an imaginary circular area around dynamic obstacles, determines the minimum safe distance for the center of the robot to approach them. It is defined by $r_{\text{obstacle}}$ plus a safe zone distance $(d_{\text{sz}})$, yielding the obstacle safe zone radius ($r_{osz}$). The safe zone distance must be equal or greater than $r_{robot}/2$, balancing collision risk and travel efficiency. Smaller distances increase collision risk but may shorten the total path, while larger distances enhance safety but might extend travel time. The most critical obstacle poses the highest risk, requiring immediate action. Intermediate targets are sub-targets the robot aims to reach before the main target, often at the intersection of tangent lines drawn from the robot's position to the obstacle's safe zone. The estimated obstacle position refers to the predicted future position of dynamic obstacles, while the collision point marks where their safe zone intersects with the robot's path. As depicted in Fig. 2, the critical area is divided into two fields: region I and region II. This division is utilized in the algorithm, which will be further explained in the methodology section. The creation of these regions proceeds as follows: first, the world frame is translated to the robot's frame without any rotation. Then, the region is labeled as region II if it lies within the symmetric quarter where the target is located, while the remaining areas are labeled as region I. Frames represent the environmental and robot conditions at specific time intervals.

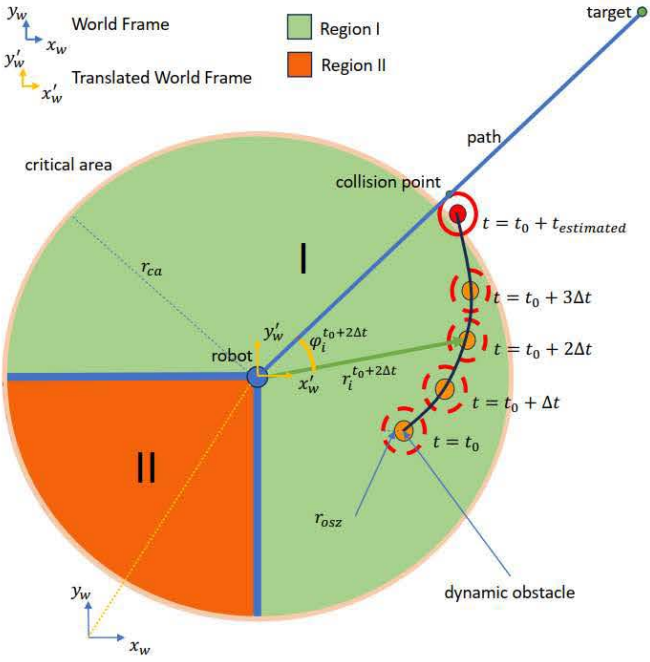

Fig. 2. Visualisation of key terms and extrapolation explanation

In Fig. 2, an angle $\phi$ and distance $r$ are also depicted, which are data provided by onboard sensors or external helper robots such as UAVs. These data consist of the angle $\phi$ and distance $r$ for each dynamic obstacle detected in the critical area. At each frame, the gathered data is kept in the robot for dynamic obstacles within the critical area, but only the last four data are stored. This limitation is because the algorithm utilizes third-order extrapolation to estimate the future position of dynamic obstacles, and additional storage would be redundant. $[r_i^t, \phi_i^t]$ represents the relative position of the $i^{th}$ dynamic obstacle with respect to the robot at time $t$, denoted by distance $r$ and angle $\phi$.

The algorithm basically consists of two stages: determination of the most critical obstacle and reaction decision against the most critical dynamic obstacle to avoid collision.

*1) Determination of the Most Critical Obstacle:* There are 4 steps for determining the most critical dynamic obstacle. Firstly, the obstacle needs to be detected in the critical area. This means that at time $t$, $r_i^t$, needs to be smaller or equal to $r_{ca} - r_{obstacle_i}$. When any obstacle is detected in the critical area, safe zones are drawn around them. After this process, all the operations are done by considering the obstacle with safe zones. When any obstacle is detected within the critical area, the robot stores their positions. These positions can be represented as follows:

$$\left\{(r_i^{t_0+k\Delta t}, \phi_i^{t_0+k\Delta t})\right\}$$

where $k \in \{0,1,2,3\}$ and $t_0$ is the first time the obstacle is detected in the critical area. If an obstacle is detected more than four times, $t_0$ corresponds to the earliest of those four instances. In other words, $t_0 + 3\Delta t$ means current time.

Once the last four positions are stored, the subsequent step involves listing dynamic obstacles where the distance between them and the robot decreases. If $r_i^{t_0+3\Delta t} - r_i^{t_0+2\Delta t}$ is smaller than zero, it means the $i^{th}$ obstacle is approaching the robot.

The third step is collecting the obstacles approaching the path, which is the line segment from the current position of the robot to the target point. This can be done by calculating the shortest distance of the obstacles to the path.

The fourth step is to label the obstacle nearest to the robot from the latest kept list as the most critical obstacle at the current time, which means the smallest $r_i^{t_0+3\Delta t}$. Reactions and maneuvers will be executed by considering this obstacle first.

*2) Reaction Decision Against the Most Critical Dynamic Obstacle to Avoid Collision:*

*2.1) Scenario Classifications:* As shown in Fig. 3, there are 6 different scenarios in which the robot can react to the most critical dynamic obstacles. Case 1 operates independently of regions within the critical area, while Case 2 occurs when the most critical obstacle is located in region II. In all other scenarios, obstacles placed in region I. The block diagram of the algorithm can be seen in Fig. 3 and the detailed explanations of these cases are provided below:

- *Case 1:* When there is no most critical obstacle in the critical area.
- *Case 2:* When the most critical obstacle approaches the robot from region II.
- *Case 3:* When the most critical obstacle is already on the path and static.
- *Case 4:* When the most critical obstacle is already on the path and dynamic.
- *Case 5:* When the most critical obstacle is approaching the robot's path and the estimated required time of its

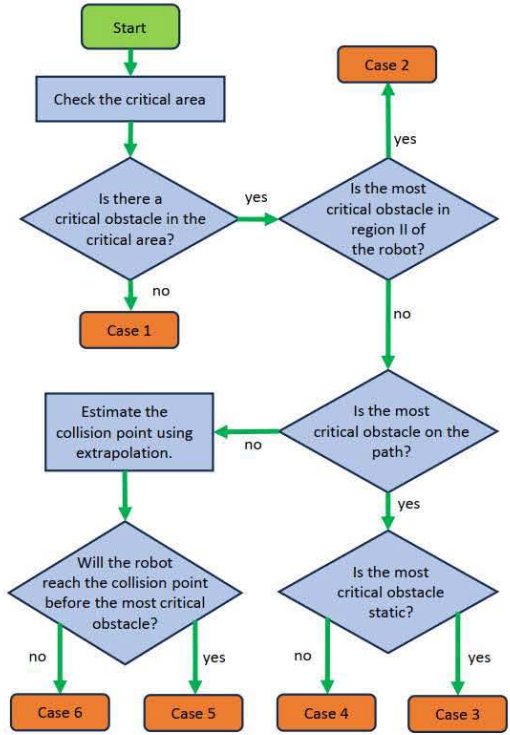

Fig. 3. Algorithm block diagram

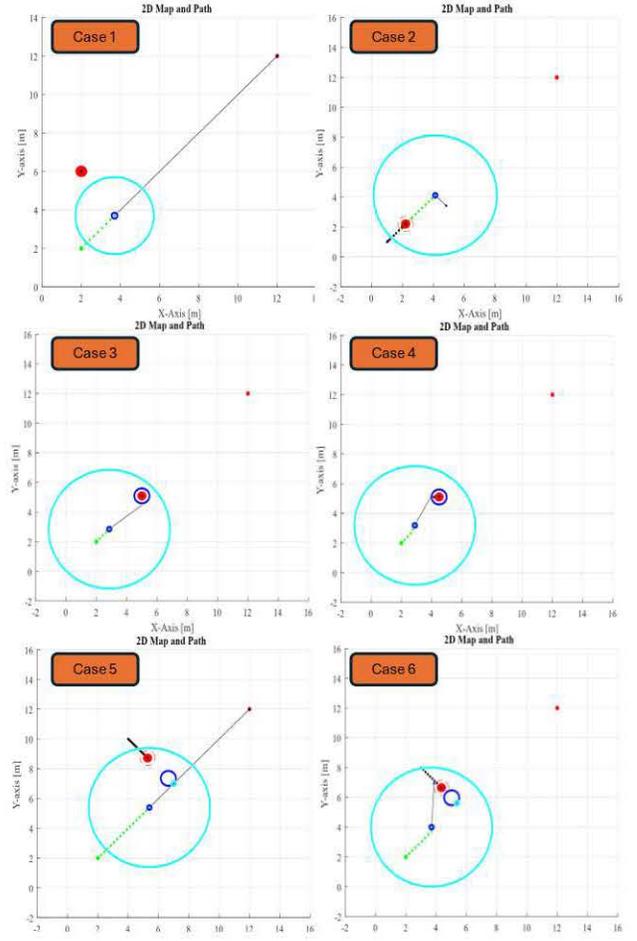

Fig. 4. Reaction against the most critical obstacle

safe zone to touch the collision point is bigger than the required time for the robot to touch the same point.
- *Case 6:* When the most critical obstacle is approaching the robot's path and the estimated required time of its safe zone to touch the collision point is smaller than the required time for the robot to touch the same point.

*2.2) Corresponding Reactions to Scenarios:* The reaction to scenarios can be seen in Fig. 4. In case 1 and case 5, the robot maintains its current direction without changing its path.

In case 2, the robot adjusts its direction by becoming perpendicular to a vector drawn from the last position of the dynamic obstacle to the robot's current position.

In case 3 and case 4, 4 tangent lines are drawn to the obstacle's safe zone circle: 2 of them from the current robot position and 2 from the target point. These 4 lines actually form 2 path solutions by intersecting. In case 3, the robot prioritizes the shortest available path, while in case 4, it selects the path from which the dynamic obstacle originated, determined by comparing $\phi_i^{t_0+2\Delta t}$ and $\phi_i^{t_0+3\Delta t}$.

For case 5 and case 6, the first step involves estimating the time required for the dynamic obstacle's safe zone to touch the collision point using extrapolation methods. Visual representation can be seen in Fig. 2. Utilizing the last 4 shortest distances and their corresponding timestamps, an estimation is made regarding when the dynamic obstacle's safe zone will intersect the path. Subsequently, using this estimated time alongside the last 4 x and y positions, further extrapolations are conducted to predict the collision position of the obstacle with the path. The time for the robot to reach the collision point is calculated considering its velocity. This calculated time is then compared with the time estimated for the dynamic obstacle, and decisions are made based on this comparison. If the time required for the robot to reach the collision point is smaller than the required time for the dynamic obstacle's safe zone to touch this point, then case 5 is determined; otherwise, case 6 is chosen.

In the case of case 5, the robot does not change its path. If it is case 6, then the robot prefers the longest path tangent to the dynamic obstacle's current position as done in case 4.

### B. Highly Dynamic Environment Creation

The positions of the dynamic obstacles with respect to time can be represented as $P_{\text{obstacle}}(x(t), y(t))$. To create a highly dynamic environment, in other words, generating higher complex trajectories and motions with acceleration, the trajectories are determined to be 2nd order polynomials.

As explained in [17], according to implicitization by Sylvester's matrix, in general, if $x$ and $y$ are polynomials of degree $p$, the corresponding algebraic curve $f(x, y) = 0$ will also have degree $p$. So the degree of $x(t)$ and $y(t)$ actually equals the degree of the position curve $P$, which is determined as 2 in the implementation. In order to get the velocity, the position vector can be differentiated with respect to time $x'(t)$ and $y'(t)$.

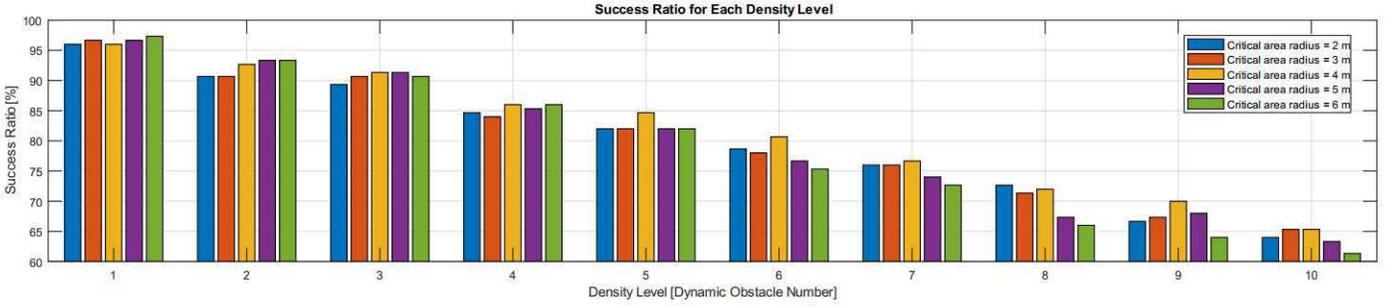

Fig. 5. Critical Area Comparison

As mentioned before, there was a velocity constraint in the problem definition. The maximum velocities of the obstacles are as maximum as the robot's. Therefore, throughout the motion, the following condition needs to be ensured.

$$V \geq \sqrt{x'(t)^2 + y'(t)^2} \quad (3)$$

In order to get the acceleration, the velocity vector can be differentiated with respect to time $x''(t)$ and $y''(t)$. There is also an acceleration limitation in the system as follows.

$$A \geq \sqrt{(x''(t))^2 + (y''(t))^2} \quad (4)$$

where A is the maximum allowable acceleration for dynamic obstacles. In other words, dynamic obstacles follow a non-linear path with non-constant speeds, which increases the complexity of the dynamic environment.

## III. RESULTS AND DISCUSSION

Simulations are carried out in MATLAB environment. The initial positions of the dynamic obstacles are assigned randomly within the area and the velocity and the acceleration constraints are taken into account. By considering previous works such as [7] and [18], the drivable area was determined to be $10 \times 10 \ m^2$. The maximum velocity was determined through research, based on various ROS robots such as HUSKY, and other models like the SR7 used in previous works [19]. As a result, the maximum velocity of the robot in simulation was set at 2 m/s, with a maximum allowed acceleration of 2 $\frac{m}{s^2}$.

To maintain the density of the drivable area, which is the number of dynamic obstacles over the drivable area, when a dynamic obstacle reaches the end of the area, its trajectory takes a symmetrical path on the map, similar to the snake game on a phone. Therefore, throughout the motion, the complexity of the environment is kept constant as initial.

Numerous scenarios were generated and algorithm was tested. Some examples of various complex scenarios can be seen in Fig. 6. The green dots symbolize the position of the UGV at each frame, and the black ones represent the positions of the dynamic obstacles. It can be inferred that if the distance between dots is long, it means the dynamic obstacle is moving faster, as it covers more distance in one time unit (which was 0.1 seconds in the simulations).

The Fig. 5 illustrates the success ratios of the local path planning algorithm in these scenarios with varying numbers of dynamic obstacles. The success ratios were analyzed across five different critical area radii, from 2 meters to 6 meters.

In low-density environments with 1 to 4 obstacles, success ratios are consistently high, above 85%, regardless of the critical area radius. This indicates the algorithm's high effectiveness in less congested environments, with minimal impact from the critical area radius. As obstacle density increases to a medium level (5 to 7 obstacles), success ratios decline but remain between 70% and 90%. Differences based on the critical area radius become more apparent, with a radius of 3 meters often performing slightly better, indicating an optimal balance between caution and agility.

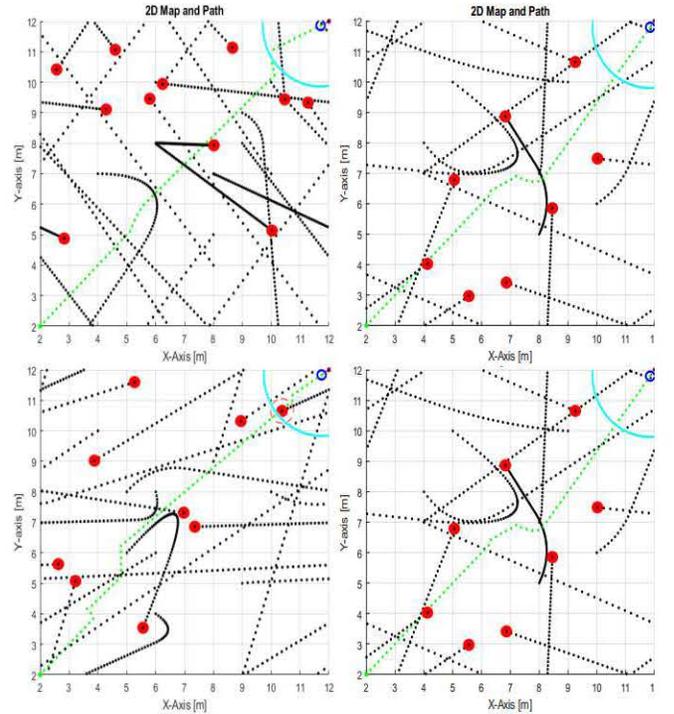

Fig. 6. Reaction against the most critical obstacle

In high-density environments with 8 to 10 obstacles, success ratios drop significantly. In complex environments, a larger critical area increases the likelihood of the most critical obstacle changing frequently, causing the algorithm to react to different obstacles at each frame. This leads to noisy and fluctuating inputs, resulting in poor navigation. Thus, a 2-meter radius performs worse in low-density environments compared to a 6-meter radius but performs better in high-density environments.

Overall, a critical area radius of 3 meters offers balanced performance across most density levels, making it a good choice.

## IV. CONCLUSION

A new approach was developed for local path planning by combining tangent-based methods and extrapolation. With the help of initial global map information obtained from UAVs or other methods, the global path is determined beforehand. The algorithm navigates a UGV between predetermined waypoints while avoiding collisions with dynamic obstacles in unstructured and complex environments. Throughout the robot's motion, detections can be performed via a helper UAV or alternative methods such as onboard sensors. The algorithm aims to enhance navigation safety and reduce travel time through these local paths when a UGV moves through scenarios that are highly complex and dynamic. The performance of the local path planning algorithm is related to the critical area, which represents the field that the robot must take into account for dynamic obstacles to ensure the algorithm functions effectively. Numerous highly dynamic environments were created, and simulations were run with different critical area radii. The success ratios were extracted and analyzed for each to determine the optimum one.

As part of future work, the hyperparameters of the algorithm, the velocity of the robot, and the radius of the critical area can be optimized using various learning methods. Additionally, the algorithm can be extended to operate in three-dimensional space, making it applicable for use in UAVs or submarines, thus supporting projects in 3D environments.